\documentclass{article}
\usepackage{arxiv}

\usepackage[utf8]{inputenc} 
\usepackage[T1]{fontenc}    
\usepackage{hyperref}       
\usepackage{url}            
\usepackage{booktabs}       
\usepackage{amsfonts}       
\usepackage{nicefrac}       
\usepackage{microtype}      
\usepackage{lipsum}		

\usepackage{multicol}

\usepackage{comment}
\usepackage{cite}
\usepackage{amsmath,amssymb,amsfonts}
\usepackage{algorithmic}
\usepackage{graphicx}
\usepackage{textcomp}
\usepackage{multirow}
\usepackage{amsmath,amsfonts,amssymb}
\usepackage{verbatim}
\usepackage{mathtools}
\usepackage{upgreek}
\usepackage{booktabs}
\usepackage[dvipsnames]{xcolor}
\usepackage{footnote}
\makesavenoteenv{tabular}
\makesavenoteenv{table}
\newcommand{\boldm}[1] {\mathversion{bold}#1\mathversion{normal}}

\DeclarePairedDelimiter\floor{\lfloor}{\rfloor}
\definecolor{fuente_inkscape}{RGB}{31,73,125}

\usepackage{amsmath}
\usepackage{url}
\usepackage{subcaption} 

\title{PreVIous: A Methodology for Prediction of Visual Inference Performance on IoT Devices}

\author{
  Delia~Velasco-Montero 
  \\ Instituto de Microelectr\'{o}nica de Sevilla \\ Universidad de Sevilla-CSIC
  \And
  Jorge~Fern\'{a}ndez-Berni 
    \\ Instituto de Microelectr\'{o}nica de Sevilla \\ Universidad de Sevilla-CSIC
  \And
 Ricardo~Carmona-Gal\'{a}n 
   \\ Instituto de Microelectr\'{o}nica de Sevilla \\ Universidad de Sevilla-CSIC
 \And
 ~\'{A}ngel~Rodr\'{i}guez-V\'{a}zquez 
   \\ Instituto de Microelectr\'{o}nica de Sevilla \\ Universidad de Sevilla-CSIC
}

\date{}
\renewcommand{\headeright}{}
\renewcommand{\undertitle}{}

\newcommand{\copyrightFooter}{
  \vfill
  \hrule
  \vspace{\the\dimexpr\baselineskip/2}\relax
  {
    \textcopyright 2020 IEEE.  Personal use of this material is permitted.  Permission from IEEE must be obtained for all other uses, in any current or future media, including reprinting/republishing this material for advertising or promotional purposes, creating new collective works, for resale or redistribution to servers or lists, or reuse of any copyrighted component of this work in other works.
  }
}

\begin{document}
\maketitle


\begin{abstract}
This paper presents PreVIous, a methodology to predict the performance of Convolutional Neural Networks (CNNs) in terms of throughput and energy consumption on vision-enabled devices for the Internet of Things. CNNs typically constitute a massive computational load for such devices, which are characterized by scarce hardware resources to be shared among multiple concurrent tasks. Therefore, it is critical to select the optimal CNN architecture for a particular hardware platform according to prescribed application requirements. However, the zoo of CNN models is already vast and rapidly growing. To facilitate a suitable selection, we introduce a prediction framework that allows to evaluate the performance of CNNs prior to their actual implementation. The proposed methodology is based on PreVIousNet, a neural network specifically designed to build accurate per-layer performance predictive models. PreVIousNet incorporates the most usual parameters found in state-of-the-art network architectures. The resulting predictive models for inference time and energy have been tested against comprehensive characterizations of seven well-known CNN models running on two different software frameworks and two different embedded platforms. To the best of our knowledge, this is the most extensive study in the literature concerning CNN performance prediction on low-power low-cost devices. The average deviation between predictions and real measurements is remarkably low, ranging from 3\% to 10\%. This means state-of-the-art modeling accuracy. As an additional asset, the fine-grained a priori analysis provided by PreVIous could also be exploited by neural architecture search engines.
\end{abstract}


\copyrightFooter


\section{Introduction}
implementation of visual processing at the edge, as opposed to the cloud, presents remarkable advantages such as reduced latency, more efficient use of bandwidth, and lessened privacy issues. These advantages are instrumental for boosting the application scenarios of the Internet-of-Things (IoT) paradigm \cite{itu12,lin17}. Edge vision algorithms must provide enough accuracy for practical deployments while making the most of the limited hardware resources available on embedded devices. Concerning accuracy, Deep Learning (DL) \cite{LeCun2015} has recently emerged as the reference framework. Deep Neural Networks (DNNs) resulting from training on massive datasets accomplish precise visual inference, greatly improving the performance of classical approaches based on hand-crafted features. However, this accuracy has a cost. The computational and memory requirements of DNNs are much more demanding than those of classical algorithms \cite{Verhelst2017}. This constitutes a challenge when it comes to incorporating DNN-based inference in the processing flow of IoT devices, which is already heavy because of other functions related to networking, power management, additional sensors, etc.  

The success of DL in enabling practical vision algorithms and unifying the procedure for a number of tasks such as image recognition, object detection, and pixel segmentation, has prompted research and development at various levels \cite{Sze2017_2}. At software level, various open-source frameworks, both from academia and industry, are accessible on the internet; each of them exploits a particular set of libraries and core system functionalities. At architectural level, new DNN models are ceaselessly reported aiming at enhancing specific aspects, e.g., higher accuracy, faster training, or shorter inference time. Regarding hardware, the pervasiveness of DNNs is forcing the inclusion of ad-hoc strategies that exploit different features of neural layers to speed up their processing. Overall, this extensive DL ecosystem is making the optimal selection of inference components according to prescribed application requirements increasingly difficult in vision-enabled IoT devices.

To assist in the aforementioned selection, we already proposed a methodology based on benchmarking and a companion figure of merit in a previous study \cite{velasco18}. However, benchmarking entails a significant and non-scalable effort because of the complexity and diversity of software libraries, toolchains, DNN models, and hardware platforms. In this paper, we describe PreVIous, a novel methodology that removes the need of comprehensive benchmarking. This methodology is based on the single characterization of PreVIousNet, a Convolutional Neural Network (CNN) specifically designed to encode most of the usual parameters in state-of-the-art DNN architectures for vision. As a result of such characterization on a particular software framework and hardware device, a prediction model is generated. This model provides a precise per-layer estimation of the expected performance for any other CNNs to be eventually run on that software-hardware combination. Seven CNN models on two different software frameworks have been thoroughly characterized to demonstrate the prediction capacity of PreVIous. Regarding hardware, this study is focused on the multi-core Central Processing Units (CPUs) available on two different low-power low-cost platforms, but the methodology could be extended to other types of devices. 

\vspace{0.5cm}

The manuscript is organized as follows. Section~\ref{sec_related} summarizes related work and sets the context to point out the contribution of PreVIous to the state of the art. An overview of CNNs is provided in Section~\ref{sec_Overview}, where their usual layers and fundamental characteristics are briefly described. Section~\ref{sec_previous} elaborates on the main elements defining PreVIous and how it has been applied in practical terms in this study. The core of PreVIous, i.e., PreVIousNet, is further described in Section~\ref{sec_NET}. The vast set of experimental results that confirm the modeling capacity of PreVIous is reported in Section~\ref{sec_results}. Finally, we draw the most relevant conclusions arising from these results in Section~\ref{sec_conclusions}.

\section{Related Work}
\label{sec_related}

As previously mentioned, the implementation of CNNs on resource-constrained devices is a remarkable challenge that has been addressed through various approaches. The common objective of all of them is to maximize throughput and inference accuracy while minimizing energy consumption. 

\textbf{Architecture design}. Several strategies have been investigated to boost inference performance of CNNs. For instance, the well-known SqueezeNet model \cite{Squeezenet_DBLP:journals/corr/IandolaMAHDK16} features a massive reduction of parameters with respect to previous models while still achieving a notable accuracy. Other architectures tailored for embedded devices have also been proposed \cite{SqueezeNext_1803-10615,squeezeDet_WuIJK16,Mobilenet_DBLP:journals/corr/HowardZCKWWAA17,MobilenetV2_1801-04381,shufflenet_ZhangZLS17,shufflenetv2_1807-11164,XNOR_NET_RastegariORF16}. These models were designed to alleviate their computational burden as a whole. In other words, they were not specifically adapted for a particular platform. Therefore, their performance significantly varies depending on the host system \cite{EVALUATING_THE_ENERGY_EFF_7723730}. A preliminary evaluation could be conducted by simply comparing the number of operations required for each network. However, this direct assessment does not usually translate into accurate values of measured performance metrics, in particular power consumption \cite{An_Analysis_of_DNN_CanzianiPC16}. 

\textbf{Performance benchmarking}. To select the optimal CNN according to performance requirements on a particular platform, recent studies have carried out systematic benchmarking on several hardware systems\cite{AI_benchmark_1810-01109,Benchmark_analysis,BenchmarkingTPU_GPU-CPU_1907-10701,velasco18}. Notwithstanding, benchmarking has a limited scope, considering the rapid evolution and massive number of CNNs, platforms, and software tools presently available.

\textbf{Architecture optimization}. Manual design of computationally efficient CNNs is time-consuming and requires experience. Several approaches for network compression have been investigated, ranging from network quantization\cite{low_precision_Cai2017DeepLW,dorefanet,BinarizedNN,IncrementalNQ} and channel pruning \cite{PrunningCNNfor_MolchanovTKAK16,NetworkTrimming_HuPTT16,CNN_simplification,ChannelPruningFor8237417} to special network implementations\cite{Lavin15b,FastTrainingFFT,Cong_Xiao_10.1007/978-3-319-11179-7_36,EFFNET_1801-06434,Kim2015CompressionOD,SpeedingupCN}. Alternatively, CNN design is currently moving from manual tuning towards automatic algorithms. Neural Architecture Search (NAS) or sequential model-based optimization algorithms have been proposed to optimize metrics such as latency, energy, and accuracy on prescribed platforms\cite{Learning_Transf_ZophVSL17,NETADAPT_1804-03230, AUTOML_1802-03494, Design_automation_1904-10616, MorphNet_1711-06798, MNASNET_1807-11626, FBNet_abs-1812-03443,SinglePathND,proxylessnas,CHAMNET_DBLP:journals/corr/abs-1812-08934,HyperPowerPA}. The incorporation of specific platform constraints to such automatic approaches involves modeling how the network architecture relates with the optimization target. Inference performance modeling is highly valuable in this context.

\textbf{Performance modeling}. 
Various reported methods aimed at an objective similar to that of PreVIous in terms of performance prediction. We can classify them according to their particular research focus:
\begin{enumerate}
    \item \textit{Hardware accelerators}: to leverage the energy efficiency of CNN accelerators such as Eyeriss \cite{Eyeriss_7738524}, an energy estimation methodology was proposed \cite{Sze_Designing_Energy_Efficient_DBLP:journals/corr/YangCS16a}, \cite{accelergy}. It relies on the energy costs of memory accesses and multiply-accumulate operations (MACs) at each level in the memory hierarchy. This methodology also includes an energy-aware pruning process for network optimization.
    
    \item \textit{GPU-based systems}: these platforms usually include power monitor tools, which facilitate modeling performance. On the basis of energy measurements extracted with such tools and taking network metrics and layer configurations as inputs, machine learning models targeting energy consumption and latency have been investigated\cite{Augur_1709-09503, NeuralPower_DBLP:journals/corr/abs-1710-05420, EnergyProfiling_1803-11151}.
    
    \item \textit{Cloud-mobile computing scenarios}: characterization of layerwise network performance enables finding the optimal DNN partition between the cloud and the mobile device while optimizing resources such as energy consumption, latency, and bandwidth\cite{EdgeIntelligenceLi2018EdgeIO, neurosurgeon, eff_comp_offloading_10.1145/3194554.3194565}.
    
   \item \textit{Automatic optimization algorithms}: modeling is especially valuable to drive the design of highly efficient convolutional and fully connected layers, thus accelerating optimization algorithms based on latency and energy metrics\cite{NETADAPT_1804-03230,proxylessnas,CHAMNET_DBLP:journals/corr/abs-1812-08934,Design_automation_1904-10616,HyperPowerPA}. Performance models leveraged by these algorithms comprise from simple look-up tables to more complex machine learning models.
   
   \item \textit{Training optimization}: to reduce training costs, some works modeled the performance of high-end GPUs during training\cite{performance_prediction_of_GPU_,PaleoAP}.
\end{enumerate}

Most of the performance modeling studies mentioned above revolve around high-end systems or energy-demanding platforms. However, IoT application scenarios clearly benefit from low-cost low-power devices. In addition, the reported models present limitations. They either focus on particular types of layers or involve an extensive benchmark. With PreVIous, we have addressed these key points. First, we have worked with inexpensive devices featuring enough computational power to perform CNN-based inference. Second, the performance models provide fine-grained per-layer information, covering many different types of layers with the characterization of a single CNN. 

All in all, the main contributions of this study are:
\begin{itemize}
\item A methodology that allows to evaluate the performance of CNN models accurately layer by layer in terms of throughput and energy consumption. The CNNs do not need to be actually run to obtain this evaluation, which is automatically generated from a previously built predictive model. \item A neural network whose characterization enables the construction of the aforementioned prediction model. This network incorporates a large variety of CNN layers and interconnections between them in order to achieve fine-grained CNN profiling. 
\item Rapid identification of layers whose execution time or energy consumption is distinctively higher than others on a particular software-hardware combination. Such layers would be the first ones to be modified by an optimization procedure or a NAS engine.
\item A broad analysis of CNNs running on different software frameworks and hardware platforms. This analysis, which has served the purpose of gauging the goodness of the proposed methodology, it is intrinsically valuable as an extensive set of measured performance metrics.
\end{itemize}

\section{Overview of CNNs}
\label{sec_Overview}

Next, as a basis for subsequent sections, we summarize fundamental aspects of CNNs, including architectural details, key network metrics, and typical implementation strategies.

\subsection{Common layers}
\label{sec_Layers}

CNN architectures usually comprise a heterogeneity of layers. In most  architectures, the first layer is fed with a 3-channel input image, and the network progressively reduces the spatial dimensions ($H\times W$) of feature maps (\textit{fmaps}), while increasing the number of channels $C$. 
Thus, each layer takes a 3D input tensor $\mathbf{I}$, performs operations that involve a set of learnt weights $\mathbf{W}$, and generates output data $\mathbf{O}$ for the next layer. Typical layers included in the majority of state-of-the-art CNNs, and covered by PreVIousNet, are briefly described below.

\begin{itemize}
\item \textbf{Convolutional} (CONV). Input data are convolved with a 4D tensor $\mathbf{W}$ composed of $N$ kernels of dimensions $k_h\times k_w \times C_{in}$. The $n$-th kernel $\mathbf{W^{(n)}}$, $n=1,...,N$ yields the $n$-th 2D output feature map in which the activations are obtained from:  
\begin{equation}
\mathbf{O}_{x,y,n} = \sum\limits_{c=1}^{C_{in}} \sum\limits_{i=1}^{k_w} \sum\limits_{j=1}^{k_h}  \mathbf{W^{(n)}}_{ i,j,c} \mathbf{I}_{x+i,y+j,c}
\label{eq:conv_layer} .
\end{equation}

This convolutional layer requires $k_h·k_w·C_{in}·H_{out}·W_{out}·N$ MACs. In  general, learnt biases, denoted by $b^{(n)}$, are also added to each output, adding $H_{out}·W_{out}·N$ MACs to the computation.  The operation of this layer is illustrated in Fig. \ref{fig_CONV}. 

\begin{figure}[b!]
\centering
\includegraphics[width=0.6\textwidth]{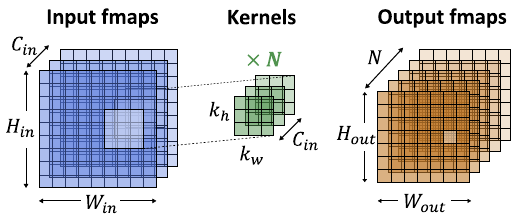}
\caption{Convolutional layers constitute the core operation of CNNs. Each kernel filter -- depicted in green -- operates on sliding local regions of the input fmaps -- \textit{receptive fields} in light blue -- to produce the corresponding output fmaps.} 
\label{fig_CONV}
\end{figure}

\item \textbf{Fully Connected} (FC). These layers are usually located at the end of the network to perform classification on the extracted feature maps. Similar to classical neural networks, the operating data are arranged in 1D vectors. A weight factor is applied to each connection between input and output activations. Additional biases can be added. Generally assuming $N_{in}$ inputs that yields $N_{out}$ outputs, a FC layer involves a computational cost of $N_{in}·N_{out}$ MACs.

\item \textbf{Pooling}. This type of layer lowers the spatial dimensions of \textit{fmaps} by applying a simple operation to each $k_h \times k_w$ patch with a stride $s$. A total of $k_h·k_w·H_{out}·W_{out}·C_{out}$ operations -- not necessarily MACs -- are performed, with $H_{out} = \floor{ \frac{H_{in}-k_h}{s}+1 }$. \textit{Maximum} and \textit{average} are the most usual functions employed to reduce dimensionality.

\item \textbf{Rectified Linear Unit} (ReLU). To introduce non-linearities between layers, various functions are applied, among which ReLU is the most popular one. It performs the simple operation of selecting the maximum between each input activation $I_{i,j,c}$ and $0$. This simplicity speeds up the calculation of non-linearities with respect to activation functions such as \textit{sigmoid} and \textit{tanh}. In addition, ReLU is more suitable for rapid training convergence \cite{MishkinSM16}.

\item \textbf{Batch-Normalization} (BN). Currently, this is the most popular normalization layer implemented in state-of-art CNNs for training acceleration. It normalizes activations on the $i$-th channel in terms of zero-mean and one-variance across the training batch \cite{BatchNorm_IoffeS15}. Two weights per channel are learnt (scale and shift), and two operations per activation are performed. 

\item \textbf{Concatenation} (Concat) of data from multiple layers, usually along the channel dimension, is convenient for merging branches in the network. For instance, this is the last layer within the \textit{Inception} module included in a number of CNNs \cite{Googlenet_DBLP:journals/corr/SzegedyLJSRAEVR14,RethinkingInception_SzegedyVISW15,Inceptionv4_DBLP:journals/corr/SzegedyIV16}. No mathematical operation is performed, only data reorganization. 

\item \textbf{Element-wise Operation} (Eltwise). This layer performs element-wise operations such as addition, product, or maximum on multiple input activations. 

\item \textbf{Scale}. This layer multiplies each input activation by a factor, thus requiring $H·W·C$ MACs. Optionally, biases can be added.

\item \textbf{Softmax} is the most notable loss function for classification tasks. It outputs a normalized probability distribution from a vector of $N$ class predictions by applying the function $softmax(I_c) = \frac{e^{I_c}}{\sum_{p=1}^{N} I_p}$.
\end{itemize}

\subsection{Intrinsic CNN Metrics}
\label{sec_CNN_Metrics}
For practical deployment of a network on resource-constrained devices, it is worth considering the following parameters: 
\begin{enumerate}
\item\textbf{Accuracy}. Once trained on a dataset for a specific application, a CNN provides a particular inference precision. In this regard, \textit{Top-N accuracy} and \textit{mean Average Precision} (mAP) are the most frequently applied metrics to evaluate classification \cite{ILSVRC15} and detection \cite{Everingham2010} tasks, respectively.

\item{\textbf{Computational Complexity ($\#OPs$)}}.  A widely used metric to measure computational complexity is the number of floating-point operations required by a network -- or, at least, those required by CONV and FC layers. 
The overall computational load can be determined by adding up the number of operations per layer -- previously detailed in Section \ref{sec_Layers}\footnote{We provide a general estimation on the minimum number of operations required for inference. Ultimately, it will be the specific interaction between hardware and software in the targeted system that will determine the actual computational complexity.}.

\item{\textbf{Model Size}}. It refers to the total amount of learnable parameters of a network. The memory footprint may preclude the execution of certain models on specific platforms. 

\item{\textbf{Memory Accesses}}. In addition to network weights, relevant activations must be kept in memory during inference.
The minimum number of basic memory operations for a layer forward-pass will be\footnote{This equation is again a plain estimation. The number of memory accesses will ultimately depend on the hardware platform -- memory word size, memory hierarchy, cache size, etc. -- and the computational strategy for each operation -- partial matrix products, data access pattern, etc.}:
\begin{equation}
\#memOPs = n(\mathbf{I}) + n(\mathbf{W}) + n(\mathbf{O})
\label{eq:memory}
\end{equation}
where $n(\mathbf{X})$ denotes the number of elements in the tensor $\mathbf{X}$. Thus, for example, $n(\mathbf{I})$ is equal to $H_{in}W_{in}C_{in}$.
\end{enumerate}

Note that these intrinsic metrics do not directly reflect actual inference performance. However, they provide a preliminary estimate of the resources required by a network. 

\subsection{Inference Metrics}
\label{sec_Inference_Metrics}
Relevant metrics concerning inference performance must be measured during forward-pass: 

\begin{enumerate}
\item{\textbf{Throughput}}. Real-time applications rely on processing images at a prescribed frame rate. CNN inference runtime limits the maximum achievable throughput for the related computer vision algorithm. 

\item{\textbf{Energy Consumption}}. Battery lifetime is one of the most critical constraints on embedded platforms. Therefore, a key parameter is the total energy demanded by the system during inference. 
\end{enumerate}

\subsection{CNN implementation strategies}
\label{sec_ImplementationStrategies}

The way in which the network architecture relates with inference metrics is highly dependent on the CNN implementation. The software libraries underlying a particular framework implement a diversity of optimization strategies to accelerate matrix multiplication according to the available hardware resources. The most commonly implemented approach is the so-called \textit{unrolled convolution}, in which convolutions are performed through image-to-column transformation (\textit{im2col}) plus General Matrix-to-Matrix Multiplication (GEMM). 
Thus, after \textit{im2col}, convolution \textit{receptive fields} are unrolled into columns, whereas filters are unrolled into rows. As a result, the convolution becomes a matrix-to-matrix product that can be highly-optimized through several libraries such as ATLAS\cite{Atlas}, OpenBLAS\cite{Goto:2008:AHM:1356052.1356053}, MKL\cite{MKL}, and cuBLAS\cite{cuBLAS}.
However, this performance optimization increases the allocated memory owing to the unrolled receptive fields. 
This memory overhead must be taken into account in Eq.~(\ref{eq:memory}), where $n(\mathbf{I})$ becomes $(k_h·k_w·C_{in})·(H_{out}·W_{out})$ for CONV layers implementing this strategy.
Other approaches that have been applied to accelerate matrix multiplication include Fast Fourier Transform (FFT)\cite{FastTrainingFFT}, Winograd\cite{Lavin15b}, and Strassen\cite{Cong_Xiao_10.1007/978-3-319-11179-7_36} algorithms.

\section{PreVIous: a Framework for Modeling and Prediction of Visual Inference Performance }
\label{sec_previous}

\subsection{General Description}
\label{sec_general_description}
The a priori evaluation of CNN performance directly on the basis of network complexity is inaccurate, even when considering only a specific hardware device \cite{An_Analysis_of_DNN_CanzianiPC16}. For the sake of more precise and multi-platform modeling, two important aspects must be stressed:
\begin{itemize}
\item Network inference involves numerous types of computational operations and data access patterns. For example, FC layers include an elevated number of weights, thus requiring a great deal of memory operations, whereas ReLU layers only perform a simple operation on activations.
\item Energy consumption is also highly dependent on the particular characteristics of the layers and how they are mapped into the underlying hardware resources. For instance, the energy cost of memory access varies up to two orders of magnitude depending on the considered level within the memory hierarchy \cite{Sze2017_2}.  
\end{itemize}

Therefore, we propose to characterize the expected performance of CNN inference through \textit{per-layer predictive models}. Not only does it make more sense according to the two aspects just mentioned, but also per-layer performance assessment is valuable for network architecture design, layer selection, and network compression.

\begin{figure}[b!]
\centering
\includegraphics[width=0.7\textwidth]{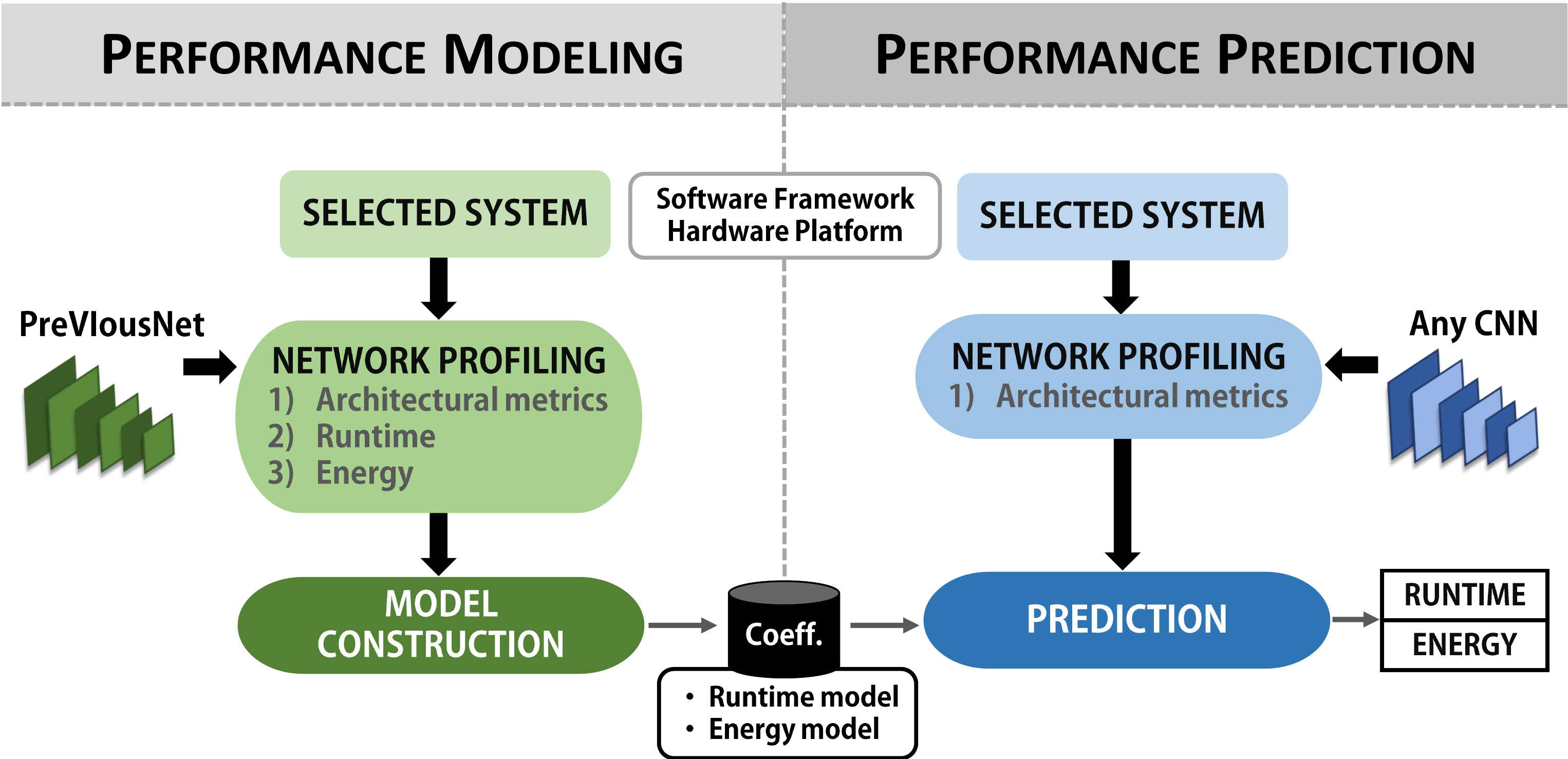}
\caption{General overview of PreVIous. It comprises two stages: 1) performance modeling, where a prediction model is constructed for the selected system through the characterization of PreVIousNet; 2) performance prediction, where the performance of any CNN of interest to be run on the selected system is accurately predicted on the basis of the previously constructed model.} 
\label{fig_PREVIOUS}
\end{figure}

Fig. \ref{fig_PREVIOUS} illustrates PreVIous. Basically, it comprises a first stage where a prediction model is constructed upon the characterization of PreVIousNet on the selected system, which is defined as a software framework implemented on a hardware platform. This one-time constructed model is able to predict, in a second stage, the performance of any other CNN to be run on such a system in terms of runtime and energy. Next, we describe this framework in detail. 

\subsection{Selected System}
\label{sec_visionSystem}

PreVIous is agnostic with respect to the software-hardware combination for modeling and prediction. In this study, we focus on two popular software frameworks deployed on two low-cost hardware platforms. The baseline combination integrates Caffe \cite{jia2014caffe} and Raspberry Pi (RPi) 3 Model B \cite{raspberry}. This embedded platform (sized $ 85\times56\times20$ mm$^3$) features a Quad Core ARM Cortex-A53 1.2 GHz CPU on a Broadcom BCM2837 System-on-Chip, 1 GB RAM LPDDR2 at 900 MHz, different network interfaces, and an external micro-SD card to provide non-volatile storage capacity. The operating system is Raspbian v9.4 Linux Kernel v4.14. All the measurements on this system were taken after booting in console mode to boost CNN inference performance and reduce energy consumption.

Once confirmed the effectiveness of PreVIous on this baseline case, we extended our analysis by changing both software and hardware in the selected system. First, we built OpenCV \cite{OpenCV} v4.0.1 on the RPi. Second, we built Caffe on a different hardware platform, namely Odroid-XU4 \cite{Odroid_XU4}. This CPU-based embedded system (sized $83\times58\times20$ mm) is more suitable for high-performance IoT applications. Its Exynos 5422 SoC implements the so-called big.LITTLE heterogeneous technology, arranging its multi-core architecture into two clusters: four ``big" Cortex-A15 2 GHz cores providing maximum computational performance and four power-efficient ``LITTLE" Cortex-A7 1.4 GHz processors. Odroid-XU4 features 2 GB LPDDR3 RAM at 933 MHz; the operating system is Ubuntu v16.04 Linux Kernel v4.14. All the measurements were also taken in console mode while running CNNs on the cluster of cores clocked at 2 GHz. This is the configuration achieving maximum throughput. 

\subsection{Network Profiling}
\label{sec_Profiling}

This step involves the extraction of three sets of measurements during the modeling stage. The first set, i.e., architectural metrics, is also extracted for network profiling during the performance prediction stage.

\begin{figure*}[b!]
\centering
\includegraphics[width=0.98\textwidth]{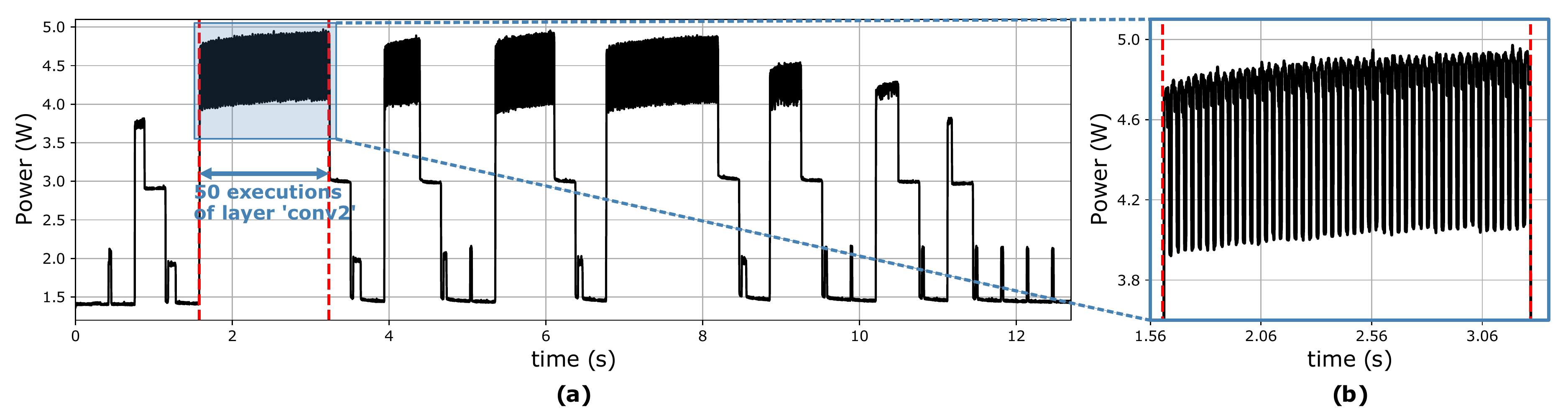}
\caption{(a) Power signal provided by a DC power analyzer when carrying out 50 consecutive executions of each layer of All-CNN-C on the combination RPi-Caffe. A time interval of 300 ms was established via software to separate each set of executions; (b) Portion of the power signal corresponding to layer `conv2' of All-CNN-C. This extracted signal is integrated and averaged over the 50 executions to obtain the energy consumption of the layer.} 
\label{fig_Power}
\end{figure*}

\begin{enumerate}
\item \textbf{Architectural Metrics}.
Simply by parsing the network definition it is possible to extract per-layer architectural parameters such as input/output dimensions ($H,W,C$), number of learnt weights, i.e., $n(\mathbf{W})$, kernel sizes ($k_h,k_w$), as well as estimates of computational ($\#OPs$) and memory ($\#memOPs$) requirements. 

\item \textbf{Time Profiling}.
Each layer composing PreVIousNet is individually run to produce per-layer runtime profiling through software functions. Specifically, we employed the \textit{time.time()} Python method to measure the elapsed time. To ensure accurate empirical measurements, 50 layer executions were averaged.

\item \textbf{Energy Profiling}.
The layers of PreVIousNet are also individually characterized in terms of energy consumption.
Some embedded platforms incorporate vendor-specific power meter tools to facilitate energy
profiling. Otherwise, a power analyzer must be connected to the power supply pins of the selected system. In our case, we connected the Keysight N6705C DC power analyzer to the power supply pins of the aforementioned hardware platforms. As an example, Fig.~\ref{fig_Power}(a) depicts the complete power profiling of All-CNN-C \cite{ALL_CNN}, the simplest among the seven networks characterized to assess the prediction capacity of PreVIous. A total of 50 executions per layer were carried out. The sampling period of the power analyzer was set to the minimum possible value, i.e., 40.96 $\upmu$s. For proper identification of the layers, a time interval of 300 ms was established via software to separate each set of 50 executions. In addition, we used the previously obtained time profiling to extract the portion of the power signal corresponding with the layer under characterization. For instance, Fig.~\ref{fig_Power}(b) shows the extracted signal for layer `conv2' of All-CNN-C. Then, the energy consumption for each layer is obtained by integrating its power signal and averaging over the 50 performed executions. 
\end{enumerate}

\subsection{Model Construction}
\label{sec_Model}

After network profiling, \textit{linear regression models} per type of layer are constructed for both runtime and energy consumption. In particular, PreVIous generates regression models for the diversity of layers listed in Section \ref{sec_Layers}, all of which are distinctively covered by PreVIousNet, as described in Section \ref{sec_NET}. Architectural metrics play the role of predictors for such models. Thus, the performance prediction stage simply consists in parsing the definition of a CNN of interest to extract its architectural metrics and apply them to the corresponding regression models for its constituent layers.

For the sake of simplicity and reducing overfitting risk, we focused on linear models. Generally speaking, a linear regression model aims at finding the best combination of a set of variables $\mathbf{x}=[x_1, x_2, ..., x_p]$ to predict the observations of a response $y$ with minimum error. Given $n$ observations of such variables and response, $\{ \mathbf{x}_i, y_i\}_{i=1}^n$, the model can be expressed as:
\begin{equation}
\mathbf{y} = \mathbf{Xw} + \mathbf{\epsilon}
\label{eq:reg_model}
\end{equation}
where $\mathbf{X}$ is the $n \times p$ matrix of predictor observations, $\mathbf{w}$ is the $p \times 1$ vector of adjusted model coefficients, $\mathbf{y}$ is the $n \times 1$ response vector, and $\mathbf{\epsilon}$ is the $n \times 1$ error vector. Note that the observations for building the model are encoded by each row $\mathbf{x}_i$ of the observation matrix $\mathbf{X}$. 


According to this notation, PreVIous produces a regression model for each type of layer in Section \ref{sec_Layers} based exclusively on architectural metrics ($\mathbf{x}$) to estimate runtime or energy  ($y$). We must point out that the observation set $\{ \mathbf{x}_i, y_i\}_{i=1}^n$ could be obtained by collecting performance data from many different CNNs. However, two undesirable facts arise from this procedure: (1) the characterization of CNNs is time-consuming and burdensome; (2) different CNNs still share many layers with similar parameters and therefore their characterization would be redundant. Alternatively, we propose a simpler and systematic approach to produce the observation set. A single characterization is needed, i.e, that of PreVIousNet, which is a new CNN that encompasses a large variety of layers and sweeps the most usual layer parameters, features, and data dimensions. Hence, the architecture design space is comprehensively covered. 

To make the most of simple linear regression models, two important points must be taken into account:

\begin{itemize}
\item \textit{Dimensionality reduction}. The higher the number of predictors, the most likely the construction of an overfitted model. Thus, only variables highly correlated with the target response $y$ must be considered for the model.
\item \textit{Model regularization}. This allows to make predictions less sensitive to a reduced set of observations. In particular, we apply standardized Ridge regularization in which the coefficients are obtained by minimizing the following expression: 
\begin{equation}
\sum\limits_{i=1}^{n} (y_i - \mathbf{x}_i \mathbf{w})^2 + \lambda \sum\limits_{j=1}^{p} w_j^2 = \\
||\mathbf{y- Xw}||^2 + \lambda ||\mathbf{w}||^2
\label{eq:Ridge}
\end{equation}
where $\lambda$ denotes the regularization tuning parameter for controlling the strength of the Ridge penalty term. We set this penalty parameter to $1$.
\end{itemize}

Considering these points, we selected $n(\mathbf{W}),\#OPs$, and $\#memOPs$ as the most meaningful predictors for building both runtime and energy per-layer regression models. We chose  these variables because they presented the highest correlation in terms of Pearson correlation coefficient when analyzing architectural parameters vs. runtime/energy in our baseline system, i.e., \textit{RPi-Caffe}.
Indeed, this inherent linear relation supports the decision of applying linear regression models rather than more complex approaches such as Support Vector Machines, neural networks, or Gaussian process regression.

\section{PreVIousNet}
\label{sec_NET}
As mentioned in previous sections, PreVIousNet is a full-custom neural network specifically conceived for modeling the performance of a variety of layers on a selected system. Therefore, it is not applicable for vision inference. Below, we first summarize common CNN architectural parameters and layer settings contemplated in PreVIousNet. Next, we describe the designed architecture. Finally, the specific configuration employed for the performance modeling stage of PreVIous is reported.

\subsection{Layer Parameters}
\label{sec_conv_settings}
Concerning convolutions, PreVIousNet includes both typical settings and special cases of CONV layers implemented in embedded CNNs:

\begin{itemize}
\item \textit{Standard CONV}. Adjustable settings include:
\begin{itemize}
\item Kernel size $(k_h, k_w)$: conventionally, an odd value is set for both dimensions. Besides, state-of-the-art CNNs feature small kernel sizes to reduce the computational load. 
\item Number of kernels $N$: to expand the channel dimension, $N>C_{in}$ kernels are normally applied.
\item Stride $s$: in case of strided convolutions, the most common value is $s=2$. 
\end{itemize}

\item \textit{Depthwise CONV}. In this type of layer, computation is saved by applying one kernel filter to each input channel. 

\item \textit{Pointwise CONV}. This non-spatial convolution uses $1 \times 1$ kernels. Two variants are possible:
\begin{itemize}
\item \textit{Bottleneck}. It reduces the computational load of subsequent layers by shrinking the channel dimension, i.e., $C_{out}<C_{in}$. 
\item \textit{General}. It increases the channel dimension without performing spatial operations. This type of layer is applied either to revert the bottleneck channel-shrinking effect\cite{ResNet_DBLP:journals/corr/HeZRS15} or to build separable convolutions\cite{Mobilenet_DBLP:journals/corr/HowardZCKWWAA17}.
\end{itemize}
\end{itemize}

Concerning other types of layers, only \textit{Pooling} layers have adjustable  $(k_h,k_w,s)$ parameters. In these layers, the maximum operation is commonly performed over $2 \times 2$ patches. In addition, some networks employ so-called global average pooling to replace memory-intensive FC layers. In this particular case, an average operation is performed on the entire input feature map. 

\begin{figure*}[b!]
\centering
\includegraphics[width=\textwidth]{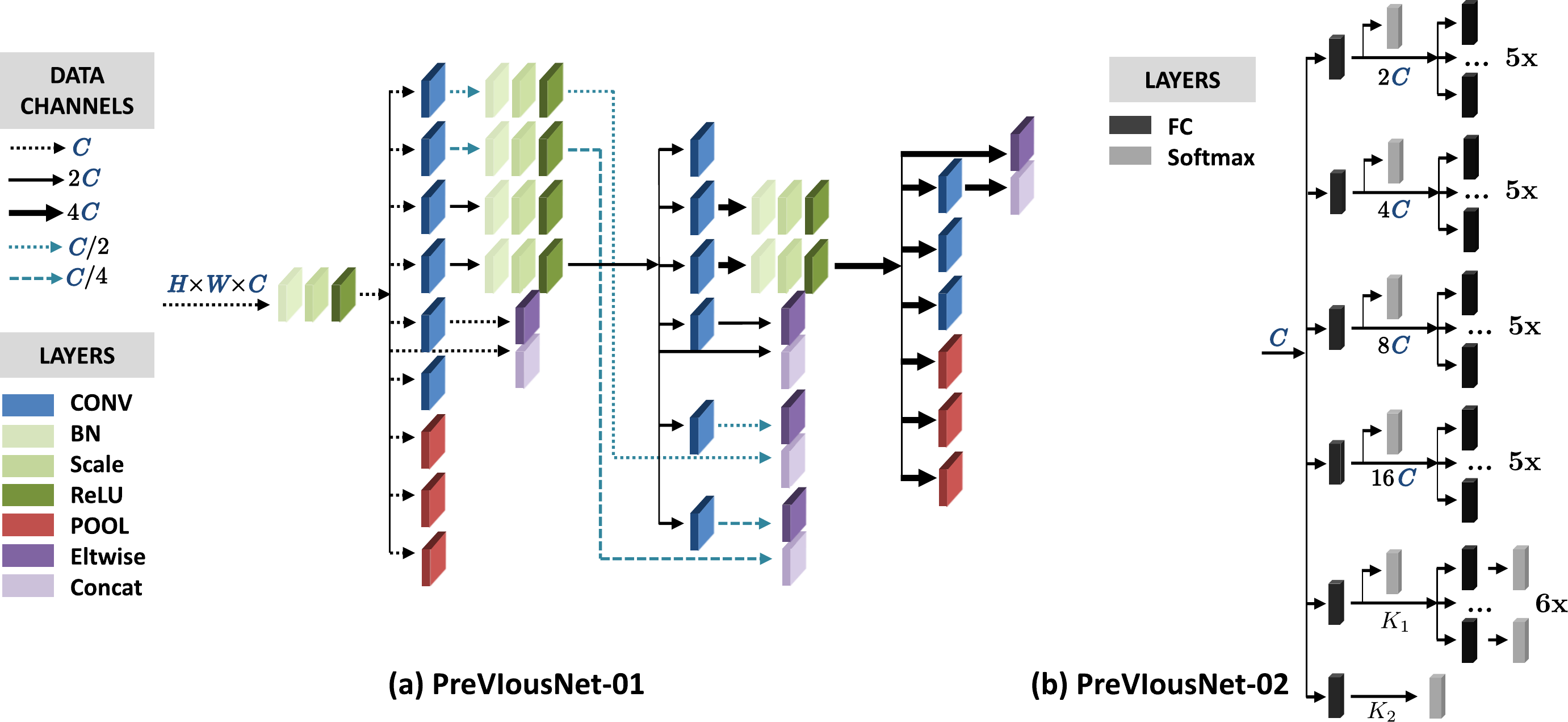}
\caption{Macroarchitecture of PreVIousNet\cite{Github_PreVIous} used in the modeling stage of PreVIous. Seven types of layers with different configurations are contemplated by PreVIousNet-01 (a), whereas \textit{FC} and \textit{Softmax} layers are covered by PreVIousNet-02 (b). The network input dimensions at the first level -- \textcolor{fuente_inkscape}{{\boldm $H$}}, \textcolor{fuente_inkscape}{{\boldm $W$}}, and \textcolor{fuente_inkscape}{{\boldm $C$}} -- are adjustable variables of PreVIousNet.} 
\label{fig_NET}
\end{figure*}

\subsection{Architecture}
\label{sec_NET_arch}
 
For each layer in a CNN, both \textit{input data dimensions} and \textit{layer parameters} determine the computational load and memory requirements, thus affecting the execution performance. This is the fact that inspired the design of the main architecture of PreVIousNet, denoted as PreVIousNet-01 in Fig.~\ref{fig_NET}(a). We aimed at covering a wide range of possibilities within the architecture design space. In this regard, note that: 
\begin{enumerate}
\item Data dimensions and computational load progressively increase as the network goes deeper -- i.e., moving rightwards in Fig.~\ref{fig_NET}(a) through the \textit{levels} of the network. The network input dimensions at the first level -- \textcolor{fuente_inkscape}{{\boldm $H$}}, \textcolor{fuente_inkscape}{{\boldm $W$}}, and \textcolor{fuente_inkscape}{{\boldm $C$}} -- are adjustable variables of PreVIousNet. 
\item Various layers and parameters are contemplated in parallel branches inserted at each level -- vertically displayed in Fig.~\ref{fig_NET}(a). 
\end{enumerate}

Based on these characteristics, PreVIousNet-01 was designed as follows. At each level, a cluster of parallel \textit{CONV} layers encompasses the aforementioned strategies: standard convolutions, pointwise, depthwise, and bottlenecks. Concerning activation layers  (\textit{BN}, \textit{Scale}, \textit{ReLU}), they perform their operation on diversely shaped intermediate \textit{fmaps} of the network -- i.e., at different levels and network branches. Likewise, \textit{Pooling} layers with varied configurations are introduced at different levels. Finally, \textit{Eltwise} and \textit{Concat} layers operate on equally sized pairs of tensors coming from previous branches of the network. Overall, PreVIousNet-01 comprises 52 layers: 15 \textit{CONV}, 7 \textit{BN}, 7 \textit{Scale}, 7 \textit{ReLU}, 6 \textit{Pooling}, 5 \textit{Eltwise}, and 5 \textit{Concat}.

Note that \textit{FC} and \textit{Softmax} layers are not included in PreVIousNet-01. These layers deal with a special case of data structure: 1D vectors instead of 3D tensors. An additional observation is that FC layers consume a notable amount of memory. Therefore, these layers are characterized through a different architecture, i.e., PreVIousNet-02, depicted in Fig.~\ref{fig_NET}(b). Thus, PreVIousNet comprises two compact specialized networks. In PreVIousNet-02, various clusters of parallel FC layers deal with input and output vectors featuring diverse sizes. The network input is a $1 \times 1 \times$ \textcolor{fuente_inkscape}{{\boldm $C$}} vector, where \textcolor{fuente_inkscape}{{\boldm $C$}} can be adjusted. Then, various sizes of data are processed by the network layers: either resulting from applying an expansion factor to the input ($2$\textcolor{fuente_inkscape}{{\boldm $C$}}, $4$\textcolor{fuente_inkscape}{{\boldm $C$}}, etc.), or using customized vector lengths $K_i$. For instance, $K_1=10$ and $K_2=1000$ can be used as they are common output sizes in classification networks trained on ImageNet\cite{ImageNet_5206848}, CIFAR\cite{CIFARdataset}, and MNIST \cite{MNISTdataset}. Similarly, \textit{Softmax} layers operate on vectors with varied dimensions. As a whole, PreVIousNet-02 includes 44 layers: 32 \textit{FC} and 12 \textit{Softmax}.

PreVIousNet is publicly available in Caffe format \cite{Github_PreVIous}. Further details about the network can be observed from this model definition. For instance, although no weight file is provided, loading the network in Caffe will automatically initialize the weights according to the ``MSRA" initialization scheme \cite{Kaiming_HeZR015}. Remarkably, the proposed architecture is not unique. It can be adjusted according to the specificity of the networks to be characterized during the performance prediction stage of PreVIous.

\subsection{Network Configuration}
\label{sec_NET_conf}
The input size of PreVIousNet-01 (\textcolor{fuente_inkscape}{{\boldm $H$}}$\times$\textcolor{fuente_inkscape}{{\boldm $W$}}$\times$\textcolor{fuente_inkscape}{{\boldm $C$}}) can be properly set according to the most common tensor sizes handled by CNNs. Let us consider SqueezeNet \cite{Squeezenet_DBLP:journals/corr/IandolaMAHDK16} as an example of embedded CNN. In this network, the number of input channels ranges from $3$ to $512$, whereas height and width of \textit{fmaps} decrease following the sequence $227,113,56,28,14$. According to this example, a characterization of PreVIousNet-01 with varied input tensor sizes is required to collect as much information as possible to build accurate prediction models. In our experiments, we empirically set the following four input dimensions: (1) $56 \times 56 \times 32$, (2) $28 \times 28 \times 64$,  (3) $14 \times 14 \times 64$, and (4) $7 \times 7 \times 64$.
Thus, we are particularly sampling\footnote{Note that in PreVIousNet-01, the number of channels increases over network levels, whereas the input \textit{fmap} resolution remains constant. The motivation is to build a simplified network to be evaluated under different configurations.} CONV layers with $H_{in}=W_{in}=\{56,28,14,7\}$ and $C_{in}=\{32,64,128,256\}$. Out of these ranges, the predicted performance values must be extrapolated from the corresponding regression models. However, this extrapolation is precise, as will be shown in Section~\ref{sec_results}.

Concerning PreVIousNet-02, we specifically run this network with an input vector sized $1 \times 1 \times 256$. Consequently, the following common vector lengths are considered $C_{in}=\{256,512,1024,2048,4096\}$, plus customized values $\{K_1,K_2\}=\{10,1000\}$. Other input configurations could be used, according to architectures of interest or device limitations.

\section{Experimental Results}
\label{sec_results}

As a first step, we completed the \textit{performance modeling} stage of PreVIous to create the models for both runtime and energy consumption on the IoT devices described in Section~\ref{sec_visionSystem}. In particular, prediction models were built upon the performance profiling of PreVIousNet-01 and PreVIousNet-02 under the 5 configurations specified in Section~\ref{sec_NET_conf} --- four for PreVIousNet-01 and one for PreVIousNet-02. Matrix $\mathbf{X}$ in Eq.~(\ref{eq:reg_model}) was obtained for each type of layer from the architectural parameters of all the corresponding layers in the aforementioned configurations of PreVIousNet. Likewise, response vector $\mathbf{y}$ in Eq.~(\ref{eq:reg_model}) was built for both runtime and energy consumption from the corresponding profilings described in Section~\ref{sec_Profiling}. 

Then, we conducted the \textit{performance prediction} stage of PreVIous on seven popular CNNs, most of them suitable for embedded devices: AlexNet\cite{AlexNet_NIPS2012_4824}, All-CNN-C\cite{ALL_CNN}, MobileNet\cite{Mobilenet_DBLP:journals/corr/HowardZCKWWAA17}, ResNet-18\cite{ResNet_DBLP:journals/corr/HeZRS15}, SimpleNet\cite{SimpleNet_DBLP:journals/corr/HasanPourRVS16}, SqueezeNet\cite{Squeezenet_DBLP:journals/corr/IandolaMAHDK16}, and Tiny YOLO\cite{YOLO9000v2_RedmonF16}. These networks were trained on ImageNet dataset\cite{ImageNet_5206848} for 1000-category classification, except for All-CNN-C and SimpleNet, which perform classification on CIFAR-10 and CIFAR-100 \cite{CIFARdataset}, respectively, and Tiny YOLO, trained on COCO dataset \cite{COCO_LinMBHPRDZ14} for object detection. As a whole, 399 CNN layers were assessed in this extensive study.

\subsection{Layerwise Predictions}
\label{sec_LayerwiseResults}

\begin{figure*}[t!]
\centering
\includegraphics[width=0.98\textwidth]{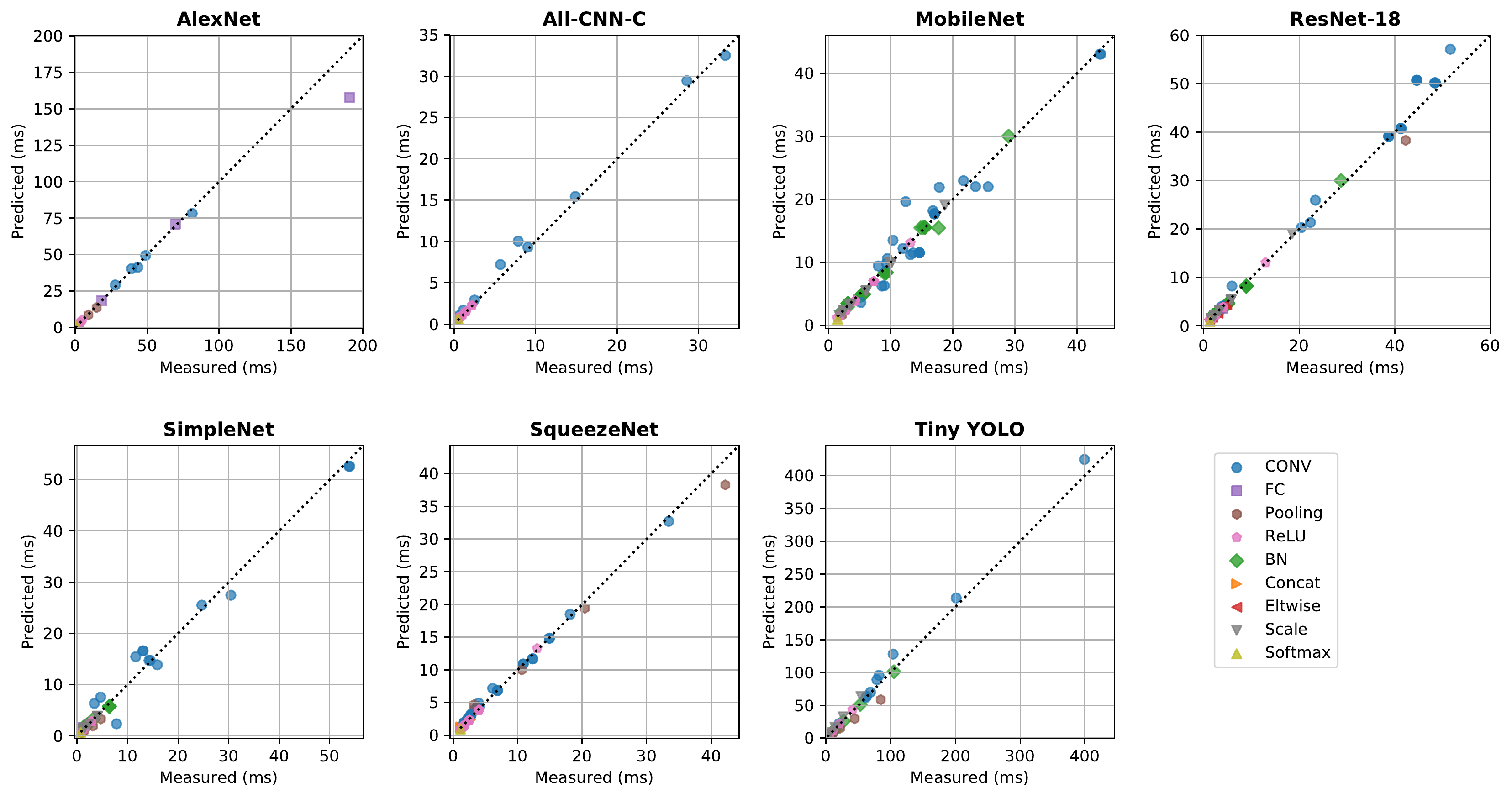}
\caption{Comparison between \textit{per-layer inference runtime} predictions (y-axis) and actual measurements (x-axis) for the baseline combination \textit{RPi-Caffe}. Similar results are obtained for \textit{RPi-OpenCV} and \textit{XU4-Caffe}. The dashed line depicts an ideal estimation in which predictions exactly match actual measurements.}
\label{fig_LayerPred_time}
\end{figure*}

To evaluate the precision of the per-layer prediction models resulting from PreVIous, we compared layerwise predictions with actual profiling measurements of the corresponding layers in all the considered CNNs. As an example, Fig.~\ref{fig_LayerPred_time} illustrates the high accuracy of the runtime predictions from PreVIous when compared to the empirical measurements in our baseline system, i.e., \textit{RPi-Caffe}. 

A summary of the results in Fig.~\ref{fig_LayerPred_time} is presented in Table~\ref{table_BasePerLayer}. Each row reports the total time obtained by adding up the runtime of all the layers composing the corresponding network, that is:
\begin{equation}
\begin{aligned}[c]
t = \sum\limits_{l=1}^{N_l} t_l
\end{aligned}
\qquad
\begin{aligned}[c]
 \hat{t} = \sum\limits_{l=1}^{N_l} \hat{t_l} 
\end{aligned}
\label{eq:Eq_SUMlayer}
\end{equation}
where $t_l$ denotes per-layer measurements, $\hat{t_l}$  denotes per-layer predictions, $N_l$ is the number of layers in the CNN, $t$ is the total measured time, and $\hat{t}$ is the total predicted time.

\begin{table}[!b]
\caption{Per-layer runtime predictions on RPi-Caffe system. Detailed profiling is shown in Fig.~\ref{fig_LayerPred_time}.}
\label{table_BasePerLayer}
\centering
\resizebox{0.48\textwidth}{!}{
\begin{tabular}{cccc}
\toprule
  & \textbf{Measured (ms)} & \textbf{Predicted (ms)} & \textbf{Error (\%)} \\
  & $t$ & $\hat{t}$ & \\
\midrule
\textbf{AlexNet}\footnote{The well-known AlexNet architecture also includes two \textit{Local 
 Response Normalization} (LRN) layers. These layers have not been contemplated in our layerwise prediction because they have been superseded by BN.} 
& 561.64 & 526.75 & -6.21\% \\
\midrule
\textbf{All-CNN-C} & 115.68 & 123.22 & 6.52\% \\
\midrule
\textbf{MobileNet} & 943.73  & 908.74 & -3.71\% \\
\midrule
\textbf{ResNet-18} & 1032.84 & 1049.10 & 1.57\% \\
\midrule
\textbf{SimpleNet} & 347.59 & 349.22 & 0.47\% \\
\midrule
\textbf{SqueezeNet} & 348.15 & 343.54 & -1.32\% \\
\midrule
\textbf{Tiny YOLO} & 1691.37 & 1740.03 & 2.88\% \\
\midrule
\multicolumn{3}{r}{\textit{\textbf{Average (absolute values)}}} & \textbf{3.24\%} \\
\bottomrule
\end{tabular} 
\begin{tabular}{l}
\end{tabular}
}
\end{table}

\begin{figure}[t!]
\centering
\includegraphics[width=0.7\textwidth]{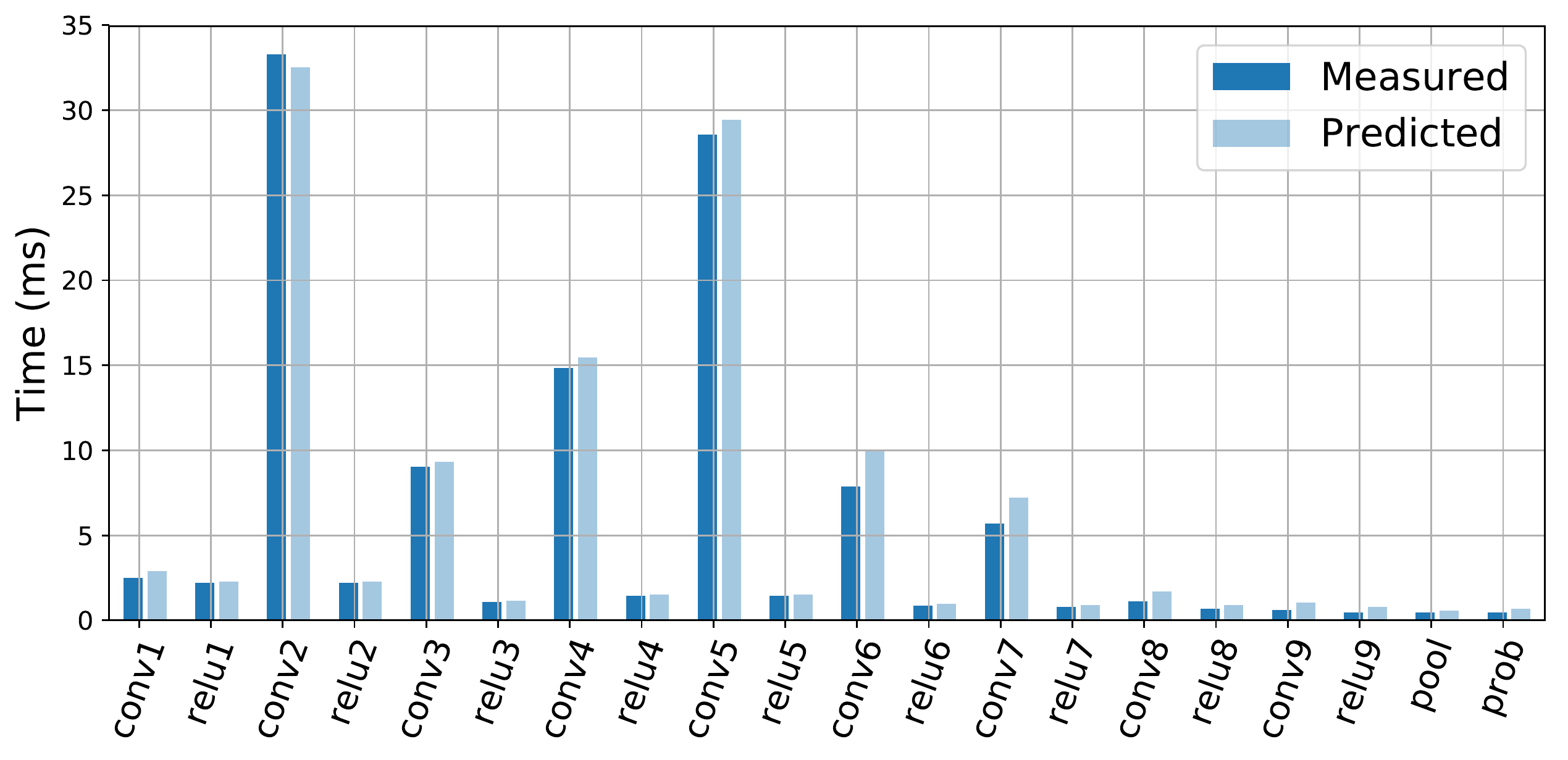}
\caption{Layerwise runtime on All-CNN-C\cite{ALL_CNN}, trained for classification on CIFAR10 dataset \cite{CIFARdataset}. Network profiling measurements were taken on the baseline combination \textit{RPi-Caffe}}
\label{fig_LayerPred_example}
\end{figure}

As an example of fine-grained performance assessment of a neural network, Fig.~\ref{fig_LayerPred_example} shows the time actually required to complete each layer of All-CNN-C\cite{ALL_CNN} in \textit{RPi-Caffe} compared with model estimations. Note that the prediction model from PreVIous correctly identifies `conv2' and `conv5' as the layers demanding the majority of the inference time. This identification is extremely useful to boost automatic optimization algorithms or NAS engines.

Finally, Table~\ref{table_perLayer} reports the average absolute error of per-layer performance prediction from PreVIous for the seven considered CNNs on the three studied systems. Remarkably, high prediction accuracy is maintained in all the cases for both runtime and energy, thereby validating PreVIous as a general framework for CNN performance modeling. 

\begin{table}[!h]
\caption{Average per-layer absolute prediction error of PreVIous for the seven considered CNNs on the three studied systems.}
\label{table_perLayer}
\centering
\begin{tabular}{lcc}
 \toprule
  & \textbf{Runtime} &  \textbf{Energy} \\
\midrule
\textbf{RPi3 - Caffe } & 3.24\% & 5.30\%  \\
\midrule
\textbf{RPi3 - OpenCV } & 4.08\% & 3.63\%  \\ 
\midrule
\textbf{XU4 - Caffe} & 3.82\% & 5.01\% \\
\bottomrule
\end{tabular}
\end{table}

\subsection{Network Predictions}
\label{sec_CNNResults}
\begin{figure}
  \begin{subfigure}[b]{0.48\textwidth}
    \includegraphics[width=\textwidth]{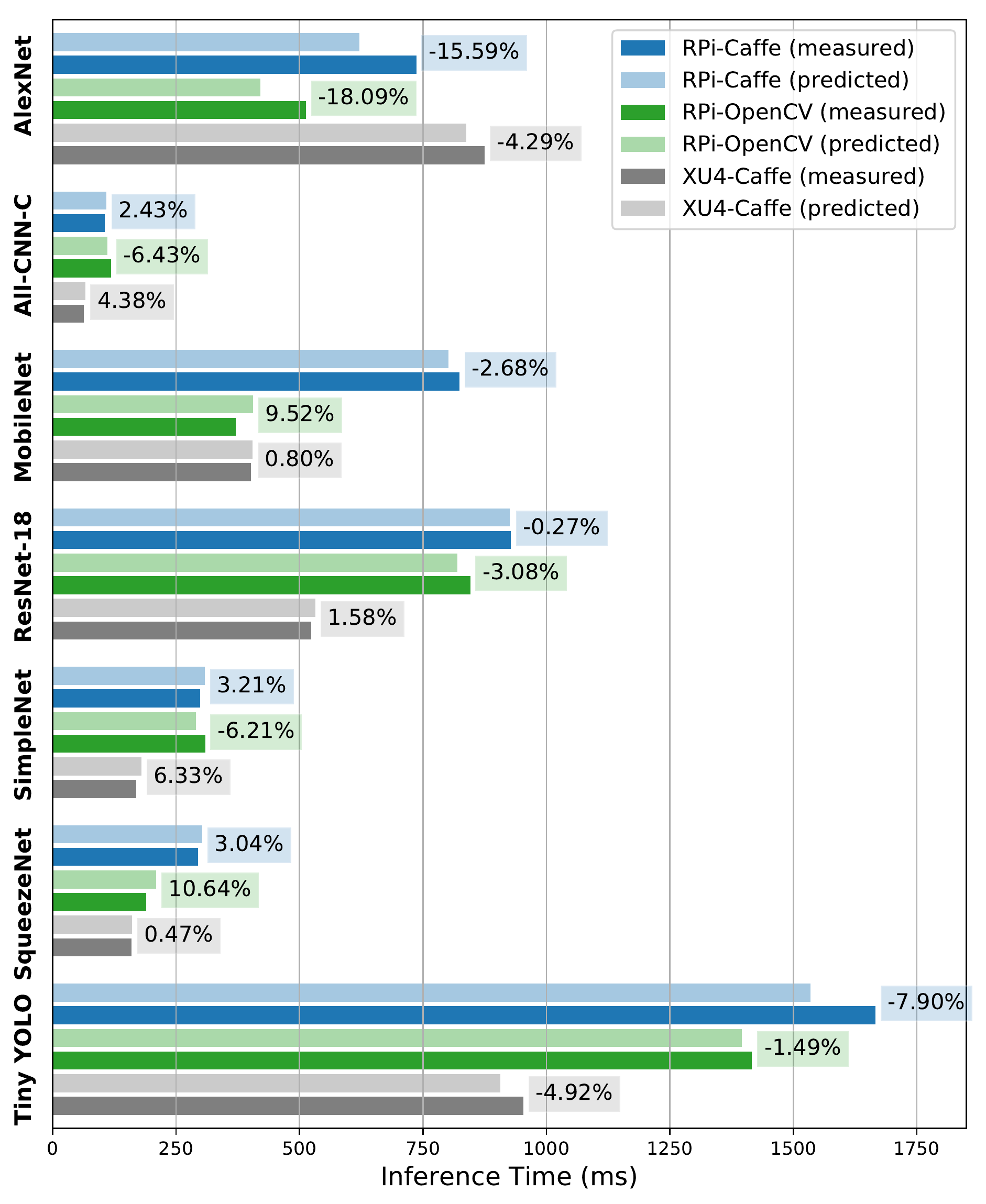}
    \caption{\textit{Network runtime predictions} vs. actual measurements for all the assessed combinations of CNN model, hardware platform, and software framework.}
    \label{fig_CNNtimePred}
  \end{subfigure}
  \hfill
  \begin{subfigure}[b]{0.48\textwidth}
    \includegraphics[width=\textwidth]{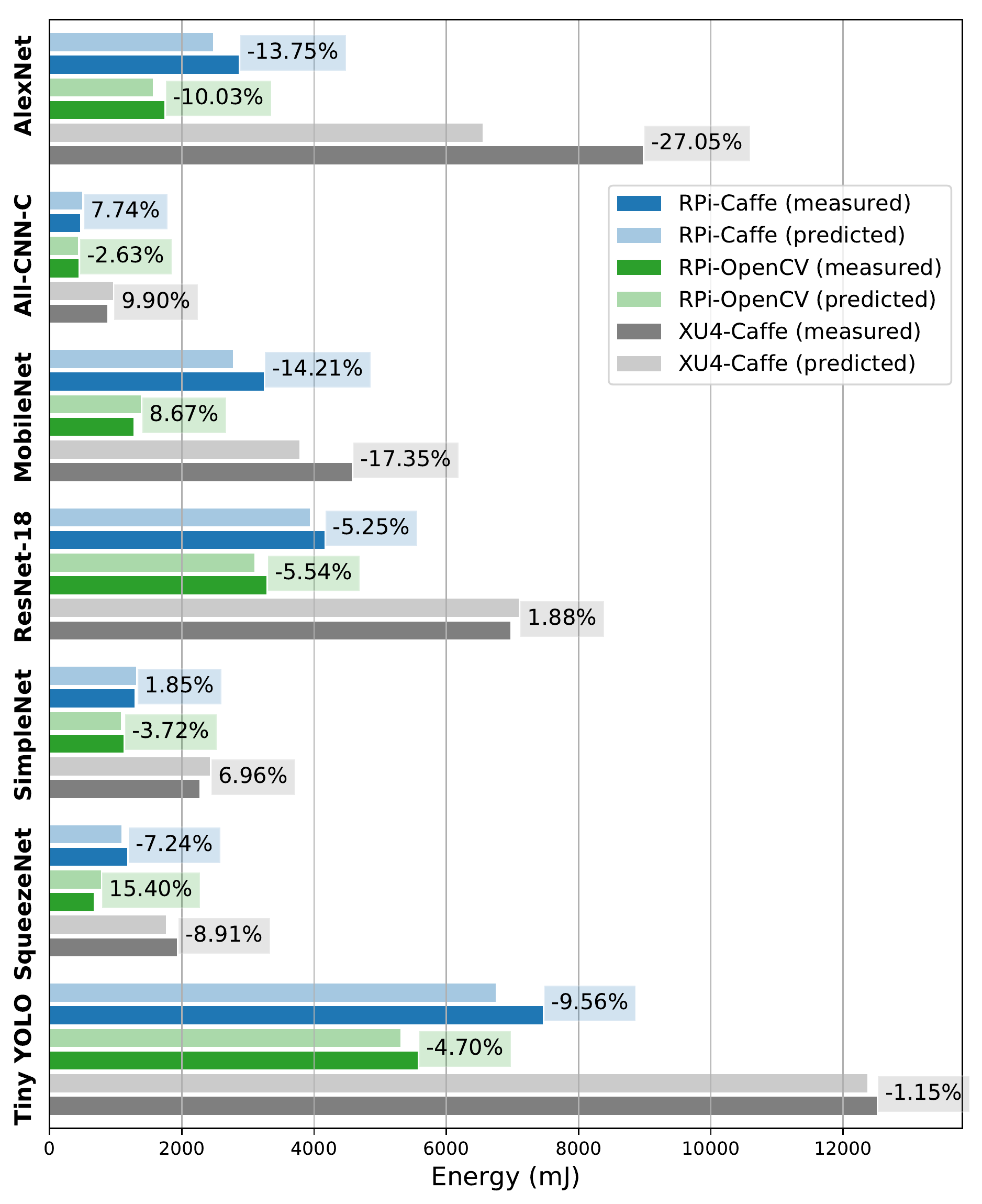}
    \caption{\textit{Network energy predictions} vs. actual measurements for all the assessed combinations of CNN model, hardware platform, and software framework.}
    \label{fig_CNNenergyPred}
  \end{subfigure}
\end{figure}

Note that Section~\ref{sec_LayerwiseResults} is focused on aggregated per-layer measurements. Indeed, a usual procedure followed in previous works on network optimization or NAS \cite{CHAMNET_DBLP:journals/corr/abs-1812-08934, FBNet_abs-1812-03443, NETADAPT_1804-03230, Design_automation_1904-10616} consists in estimating the global forward-pass performance of a network by adding up per-layer metrics, as expressed in Eq.~(\ref{eq:Eq_SUMlayer}) for runtime.
In principle, this approach should be valid given that layers are sequentially executed during CNN inference in many realizations. However, in practice, when per-layer measurements have been independently taken, their direct addition may not coincide with the actual network inference performance\cite{Augur_1709-09503, NeuralPower_DBLP:journals/corr/abs-1710-05420,eff_comp_offloading_10.1145/3194554.3194565}. This mismatch arises from aspects such as software optimizations (e.g., layer fusion or constant folding) and processor strategies (e.g., data prefetching or data re-utilization in the memory hierarchy). 
To take this fact into account, we measured the performance of the complete forward-pass of PreVIousNet in terms of runtime and energy. This forward-pass characterization allows us to write the following expression:

\begin{equation}
\hat{y} = c \sum\limits_{l=1}^{N_l} \hat{y_l}
\label{eq:Eq_CNNfactor}
\end{equation}
where $\hat{y}$ represents the total predicted runtime or energy, $\hat{y_l}$ denotes the per-layer predictions either for runtime -- $\hat{t_l}$ in Eq.~(\ref{eq:Eq_SUMlayer}) -- or energy, and $c$ is a coefficient resulting from linear regression between the direct addition of predictions and the corresponding actual measurement for the complete forward-pass of the 5 configurations of PreVIousNet on each software-hardware combination. The values of $c$ for each case are reported in Table~\ref{table_c_coeff}. Eq.~(\ref{eq:Eq_CNNfactor}) enables the comparison of  predictions from PreVIous for \textit{complete network inference} against the corresponding experimental measurements. This comparison is depicted in Figs.~\ref{fig_CNNtimePred} and~\ref{fig_CNNenergyPred} for runtime and energy, respectively\footnote{For AlexNet, profiling measurements of the deprecated LRN layers were added in the summation in Eq.~(\ref{eq:Eq_CNNfactor}) for the sake of fair comparison.}. PreVIous also provides good estimates of complete forward-pass inference, with deviations below 10\% in 18 out of 21 studied cases for runtime and 15 out of 21 cases for energy.

\begin{table}[!h]
\caption{Value of the empirical coefficient $c$ in Eq.~(\ref{eq:Eq_CNNfactor}) for the three studied systems.}
\label{table_c_coeff}
\centering
\begin{tabular}{lcc}
 \toprule
  & \textbf{Runtime} &  \textbf{Energy} \\
\midrule
\textbf{RPi3 - Caffe } & 0.88 & 1.08 \\
\midrule
\textbf{RPi3 - OpenCV } & 0.85 & 0.89 \\ 
\midrule
\textbf{XU4 - Caffe} & 0.93 & 1.09 \\
\bottomrule
\end{tabular}
\end{table}

Additionally, we further analyzed the less accurate cases such as the predictions on AlexNet -- specially for energy modeling on Odroid XU4. Regarding this network, both runtime and energy consumption were underestimated by PreVIous in all cases, as shown by the negative errors reported in  Figs.~\ref{fig_CNNtimePred} and~\ref{fig_CNNenergyPred}. 
The reason of this behavior lies in the divergence between per-layer and complete-CNN inference performance. This divergence is due to the fact that, as reflected in Table~\ref{table_c_coeff}, we apply a single value for the $c$ term in Eq.~(\ref{eq:Eq_CNNfactor}) per hardware-software combination for the sake of simplicity and generalization. For instance, in the worst prediction case -- AlexNet\footnote{Note that the well-known AlexNet model can be considered the starting point in the evolution leading to current CNN models. Therefore, its architecture somehow differs from modern ones, featuring deprecated layers and an elevated amount of learnt parameters.} energy estimation on \textit{XU4-Caffe} --, the ratio between per-layer and complete-CNN forward-pass measurements is 1.60, which is notably higher than the value of $c$ resulting from linear regression for the combination \textit{XU4-Caffe}, i.e., 1.09. The same situation was identified in \textit{RPi-Caffe} and \textit{XU4-Caffe} for MobileNet energy consumption. By contrast, per-layer predictions of PreVIous were certainly accurate, with absolute network errors below 5\% in the vast majority of the 42 studied cases -- see summaries in  Table~\ref{table_perLayer}.

\begin{table}[b!]
\caption{Comparison of CNN modeling accuracy between this study and related works in the literature. MAPE stands for Mean Absolute Percentage Error associated with complete network inference.}
\label{table_relatedWorkALL}
\centering
\resizebox{\textwidth}{!}{
\begin{tabular}{c|llccc}
\multicolumn{5}{c}{\textbf{Network Runtime}} \\
\toprule
\textbf{Ref.} & \textbf{System} & \textbf{CNNs} & \textbf{MAPE} & \\
\toprule
\multirow{4}{*}{\textbf{\cite{Augur_1709-09503}\textsuperscript{\textdagger}}} & TK1 CPU -- Caffe & \multirow{2}{*}{NIN, VGG19M} & 4.71\% &  \\
 & TK1 GPU -- Caffe  &  & 23.70\% &  \\
& TX1 CPU -- Caffe & \multirow{2}{*}{NIN, VGG19M, SqueezeNet, MobileNet} & 39.91\% \\ 
 & TX1 GPU -- Caffe  &  & 31.51\% \\
\midrule
\multirow{3}{*}{\textbf{\cite{NeuralPower_DBLP:journals/corr/abs-1710-05420}\textsuperscript{\textdaggerdbl}}} & Titan X GPU -- TensorFlow & VGG16, AlexNet, NIN, Overfeat, CIFAR10-6conv & 7.96\% &  \\
 &  GTX1070 GPU -- TensorFlow & \multirow{2}{*}{AlexNet, NIN} & 12.32\% \\
 & GTX1070 GPU -- Caffe &  & 16.17\% &  \\
\midrule
\multirow{3}{*}{\textbf{This study}\textsuperscript{\maltese}}  & RPi3 CPU -- Caffe & \multirow{3}{*}{AlexNet, All-CNN-C, MobileNet, ResNet-18, SimpleNet, SqueezeNet, Tiny YOLO} & 5.02\% &  \\
  & RPi3 CPU -- OpenCV & & 7.92\% &  \\
  & XU4 CPU -- Caffe & & 3.25\% &  \\
\bottomrule
\multicolumn{5}{c}{} \\
\multicolumn{5}{c}{\textbf{Network Energy}} \\
\toprule
\textbf{Ref.} & \textbf{System} & \textbf{CNNs} & \textbf{MAPE} & \\
\toprule
\multirow{2}{*}{\textbf{\cite{Augur_1709-09503}\textsuperscript{\textdagger}}} & TX1 CPU -- Caffe & \multirow{2}{*}{NIN, VGG19M, SqueezeNet, MobileNet} & 39.08\%\\
 & TX1 GPU -- Caffe  &  &  15.30\%\\
\midrule
\multirow{3}{*}{\textbf{\cite{NeuralPower_DBLP:journals/corr/abs-1710-05420}\textsuperscript{\textdaggerdbl}}} & Titan X GPU -- TensorFlow & VGG16, AlexNet, NIN, Overfeat, CIFAR10-6conv & 2.25\% \\
 &  GTX1070 GPU -- TensorFlow & \multirow{2}{*}{AlexNet, NIN} & 8.40\% \\
 & GTX1070 GPU -- Caffe &  & 21.99\% \\
\midrule
\multirow{2}{*}{\textbf{\cite{EnergyProfiling_1803-11151}\textsuperscript{\S}}} & \multirow{2}{*}{TX1 CPU -- Caffe} & AlexNet, ResNet-50, SqueezeNet, GoogLeNet, SqueezeNetRes, &  \multirow{2}{*}{12.26\%} 
\\ && VGG-small, Places-CNDS-8s, All-CNN-C,  Inception-BN, MobileNet  \\
\midrule
\multirow{3}{*}{\textbf{This study}\textsuperscript{\maltese}}  & RPi3 CPU -- Caffe & \multirow{3}{*}{AlexNet, All-CNN-C, MobileNet, ResNet-18, SimpleNet, SqueezeNet, Tiny YOLO} & 8.52\% &  \\
 & RPi3 CPU -- OpenCV & & 7.24\% &  \\
 & XU4 CPU -- Caffe & & 10.46\% &  \\
\bottomrule
\multicolumn{5}{l}{} \\
\multicolumn{5}{l}{Types of assessed layers: \textsuperscript{\textdagger} CONV; \textsuperscript{\textdaggerdbl} CONV, FC, and Pooling; \textsuperscript{\S} CONV; \textsuperscript{\maltese} CONV, FC, Pooling, ReLU, BN, Concat, Eltwise, Scale, and Softmax.}
\end{tabular}
}
\end{table}

\subsection{Discussion}
Some key points must be stressed about the results presented in this section: 

\begin{enumerate}

    \item We have intentionally assessed a diversity of network layers, as opposed to previous approaches, mostly focused on CONV layers. Indeed, Fig. \ref{fig_LayerPred_time} highlights the non-neglibible -- even dominant in some cases -- contribution of certain layers, e.g., Pooling in SqueezeNet or BN in MobileNet, to the total inference time. This proves the importance of their consideration for performance modeling \cite{DeepReBirth_1708-04728,shufflenetv2_1807-11164,Augur_1709-09503}.
    \item The proposed methodology has been validated on various systems suitable for IoT applications. This verification includes the process of model construction and the prediction capacity of PreVIous. The remarkable aspect here is that only five systematic network characterizations -- i.e., the considered configurations of PreVIousNet -- suffice to build accurate prediction models for a particular system. 
    
    \item Our study is, to the best of our knowledge, the most comprehensive in the literature in joint terms of number of CNNs, types of layers, performance metrics, and hardware-software combinations. The prediction accuracy is also, in global terms, the highest among similar reported works. Table \ref{table_relatedWorkALL} presents a comparison of our study vs. such similar works\footnote{Among the related studies described in Sec.~\ref{sec_related}, this table contains those which addressed global CNN performance modeling –- as opposed to single-layer characterization -– and reported numerical prediction results.}. Note that most of them made use of high-end GPUs; in our case, we focused on low-cost low-power small-sized IoT devices. The last column summarizes the prediction accuracy for each case in terms of Mean Absolute Percentage Error (MAPE), which is defined over the considered CNNs as the average of the absolute value of the difference between the complete network prediction and the corresponding actual measurement divided by the measurement. Thus, the values of this column in our case are the average of the absolute values in Figs.~\ref{fig_CNNtimePred} and~\ref{fig_CNNenergyPred} for each selected system. Concerning related works, we calculated the MAPE according to the individual errors reported for each characterized CNN. Note that we have covered a much wider spectrum of layers than the other studies, i.e., 9 types of layers vs. 3 types at most. This is the basis for achieving better predictions over a larger set of CNNs following a common procedure for both runtime and energy.  
\end{enumerate}

\section{Conclusions}
\label{sec_conclusions}

This study demonstrates that it is possible to predict the performance of CNNs on embedded vision devices with high accuracy through a simple procedure. Taking into account the growing and ever-changing zoo of CNN models, such a priori prediction is key for rapid exploration and optimal implementation of visual inference. The utility of the proposed methodology, i.e., PreVIous, is two-fold. First, fine-grained layer performance prediction facilitates network architecture design and optimization. Second, network performance estimation can assist in CNN selection to fulfill prescribed IoT requirements such as latency and battery lifetime. 

Simplicity is indeed a major asset of PreVIous. Only the characterization of a single architecture is required for performance modeling. We also make use of linear regression to reduce model complexity. In addition, the procedure does not rely on any specific measurement tool, being agnostic with respect to the selected hardware-software combination.

Future work will address the design of further versions of PreVIousNet in order to consider new types of layers or even entire building blocks. For instance, recurrent building blocks of highly optimized architectures can be characterized as a whole, e.g., the \textit{Fire} module of SqueezeNet or \textit{separable convolutions} of MobileNets. This approach can also be exploited by automatic algorithms to explore new architectures optimally adapted to specific embedded systems.

\section{Acknowledgments}
This work was supported by Spanish Government MICINN (European Region Development Fund, ERDF/FEDER) through project RTI2018-097088-B-C31, European Union H2020 MSCA through project ACHIEVE-ITN (Grant No. 765866), and by the US Office of Naval Research through Grant No. N00014-19-1-2156.

\bibliographystyle{IEEEtran} 
\bibliography{bibliography} 

\end{document}